%% file: root.tex
\documentclass[letterpaper, 10 pt, conference]{ieeeconf} 

\IEEEoverridecommandlockouts                        

                            \usepackage{tikz}
\usetikzlibrary{shapes.geometric, positioning, calc}
\usepackage{graphics}
\usepackage{graphicx}
\usepackage{subfigure} 
\usepackage{amsmath} 
\usepackage{amssymb}
\usepackage{cite}
\usepackage{tikz}
\usepackage{amsmath}
\usepackage{tabularx}
\usepackage{array}
\usepackage{booktabs}
\usepackage{bm}
\usepackage{algorithm}
\usepackage{algpseudocode}
\usepackage{bbm}
\usepackage{caption}
\PassOptionsToPackage{hyphens}{url}\usepackage{hyperref}
\usepackage{float}
\usepackage{multirow}

\usepackage{tablefootnote}

\definecolor{green}{RGB}{11,155,13}

\title{\LARGE \bf
Failing Gracefully: Mitigating Impact of Inevitable Robot Failures
}

\author{Duc M. Nguyen, Saad A. Ghani, Andrew Marshall, Allison Andreyev, Gregory J. Stein, and Xuesu Xiao
\thanks{Duc M. Nguyen, Saad A. Ghani, Gregory J. Stein, and Xuesu Xiao are with the Department of Computer Science, George Mason University. Emails: {\tt\scriptsize \{mnguy21, sghani2, gjstein, xiao\}@gmu.edu}.}%
\thanks{Andrew Marshall and Allison Andreyev are high school interns affiliated with the RobotiXX Lab, George Mason University. Emails: {\tt\scriptsize \{amarshall10003, allisonmandreyev\}@gmail.com}.}
}

\begin{document}
\maketitle
\thispagestyle{empty}
\pagestyle{empty}

\input{content/abstract.tex}

\input{content/intro.tex}
\input{content/related.tex}
\input{content/failure_mitigation.tex}
\input{content/failbench}
\input{content/method.tex}
\input{content/future.tex}
\input{content/conclusions.tex}

\section*{Acknowledgment}
This work has taken place in the RobotiXX Laboratory at George Mason University. RobotiXX research is supported by National Science Foundation (NSF, 2350352), Army Research Office (ARO, W911NF2320004, W911NF2520011), Google DeepMind (GDM), Clearpath Robotics, FrodoBots Lab, Raytheon Technologies (RTX), Tangenta, Mason Innovation Exchange (MIX), and Walmart.

\bibliographystyle{IEEEtran}
\bibliography{IEEEabrv,references}
\end{document}

%% file: content/abstract.tex
\begin{abstract}
Service robots operate in household environments shared with humans, pets, and everyday objects, where they are highly susceptible to failures such as software crashes, hardware degradation, or unpredictable interactions. While roboticists strive to minimize failures, some remain inevitable, making it critical to mitigate their potential consequences for safe and reliable deployment. This paper introduces a novel safety formulation that evaluates both the probability of impactful interactions between robots and surrounding entities during failures, and the severity of their outcomes. By quantifying the impact of failures on different entities, our approach enables robots to make informed planning decisions that balance safety with task efficiency. To support systematic evaluation, we also present \texttt{FailBench}, a MuJoCo-based simulation framework for studying robot-environment interactions under diverse failure modes, including sensing issues and actuator malfunctions. Together, our safety formulation and \texttt{FailBench} provide a foundation for developing safer and  more robust motion plans and learned policies in real-world household environments.

\end{abstract}

%% file: content/intro.tex
\section{Introduction}
\label{sec::introduction}

Household robots are expected to perform diverse tasks such as cleaning, fetch-and-deliver, and cooking. These tasks require operation in dynamic environments shared with humans, pets, and cluttered objects, where robots are highly susceptible to failures such as falling, dropping objects, or spilling liquids. Such failures may arise from hardware degradation, software bugs, or unpredictable conditions. For example, a spilled drink can short-circuit electronics and create a slippery floor that risks human injury; a dropped object may break fragile items; and a fall can injure a nearby person or pet. Ensuring safety in the presence of failures is therefore crucial for the widespread adoption of service robots in our homes.

Most existing work emphasizes failure prevention and recovery strategies~\cite{xiao2020robot, de2011minimum, zabarankin2002optimal, ono2008efficient, cornelio2024recover}, but often lacks systematic failure categorization and overlooks the \textit{consequences} of inevitable failures. In household settings, failures extend beyond incomplete tasks (e.g., getting stuck at a doorway) to hazardous events that can damage property or endanger people. 

While previous failure prevention and recovery strategies can lower the chance of failures, they do not consider the consequences if an inevitable failure occurs in the planning process. One naive way to mitigate failure impact is through inflating objects or defining ``no-go zones''. But by trimming down the workspace to stay away from everything unnecessarily limits where the robot can operate and compromise task efficiency. 

Another major bottleneck in studying such failure consequences is the scarcity of real-world data during real-world service robot deployment and the lack of a standard benchmark to evaluate how robots can mitigate failure impact. 

Without sufficient data and a benchmark, it is difficult for robots to analyze diverse failure consequences to plan motions or to  learn policies that distinguish minor and severe failures to simultaneously maintain efficiency and safety. 

To enable robots to reason about and mitigate failure consequences, we propose a novel problem formulation that explicitly accounts for the \textit{consequences} of failures during planning. Our framework evaluates potential \textit{interactions} between robot components and environmental entities when failures happen, quantifies the \textit{severity} of each interaction, and aggregates them into an overall impact measure. Achieving this requires reasoning not only about \textit{when} failures occur, but also \textit{what} they impact and \textit{how severe} the resulting consequences are.

\begin{figure}[t]
    \centering
    \includegraphics[width=0.47\textwidth]{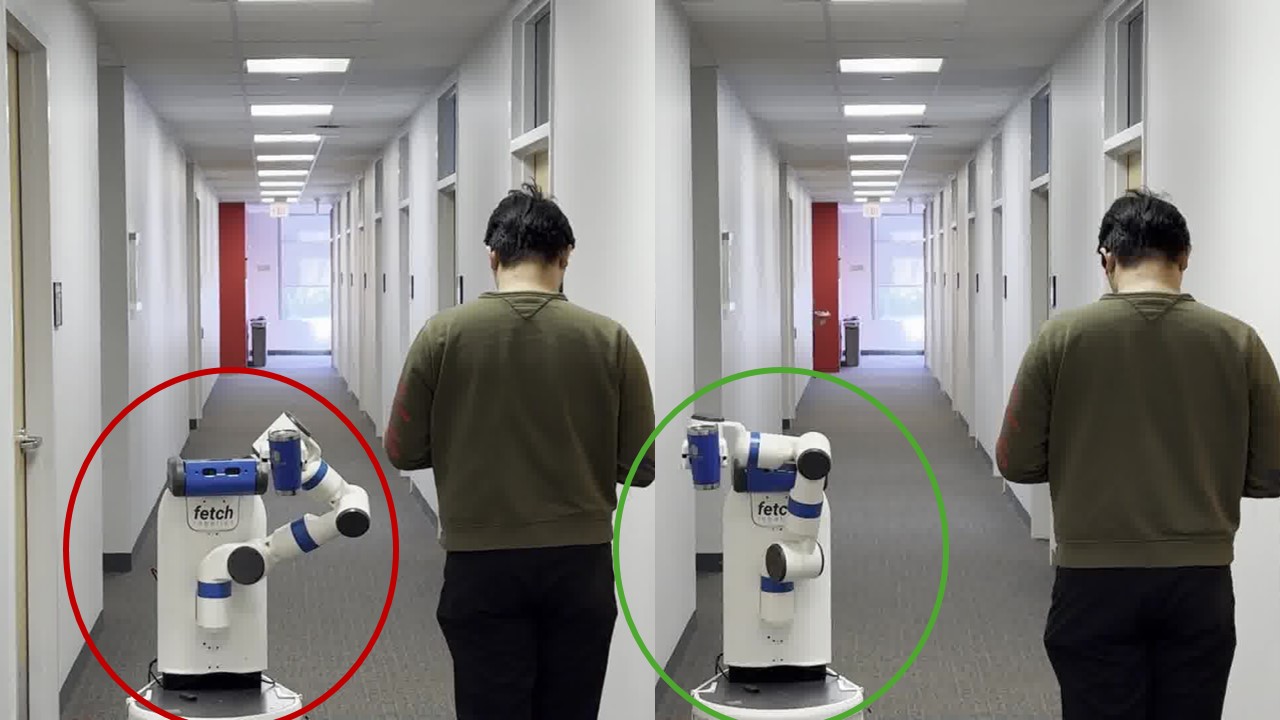} 
    \caption{Robot failure mitigation example. \textbf{Left} (red): Carrying a bottle of hot water close to a human poses a higher risk of spilling and causing injury if dropped. \textbf{Right} (green): A safer strategy positions the bottle farther away from the human to reduce potential harm in the event of failure.}
    \label{fig:single_image}
\end{figure}

For example, in Figure \ref{fig:single_image}, a robot carrying a bottle in a hallway may position the bottle at a lateral offset from humans, not only to prevent collisions, but to limit the impact on the person if the bottle is accidentally dropped.

To provide failure data and benchmark, we present \texttt{FailBench}, a MuJoCo-based simulation framework designed to study how robots interact with their environments under sudden unexpected failures such as hardware shutdowns, sensor degradation, or actuator malfunctions. \texttt{FailBench} enables systematic analysis of failure-induced behaviors, supporting the development of safer motion plans and more robust learned policies. The framework provides: (1) a curated library of scenes and objects household robots may commonly encounter, (2) a suite of Python implementations of common motion planners for both navigation and manipulation, and (3) a novel failure injector that introduces controlled perturbations to the robots and its environments to collect failure data and benchmark failure mitigation strategies. By offering a controlled yet diverse environment for benchmarking, \texttt{FailBench} facilitates rigorous evaluation of robot safety and resilience in the presence of failures, ultimately guiding safer real-world deployment.  

In summary, this paper makes the following contributions:
\begin{itemize}
    \item A novel formulation for failure impact assessment with a new safety metric, combining interaction modeling and severity metrics to quantify safety risks;    
    \item \texttt{FailBench}, a MuJoCo-based benchmark to study failure mitigation by generating diverse failure data and systematically evaluating failure impacts; and
    \item Demonstration that our new formulation can reflect failure consequences simulated in \texttt{FailBench}, paving the way toward future research into failure mitigation for service robots. 
\end{itemize}

%% file: content/related.tex
\section{Related Work}
\label{sec::related_work}

\textbf{Robot Failure} is a critical consideration for the widespread adoption of robotic systems, with prior work aiming to reduce failure likelihood and develop robust risk formulations~\cite{honig2018understanding, carlson2004follow, xiao2020robot, xiao2019explicit}. Failures have been categorized as \textit{physical} (e.g., power loss, sensor errors, communication faults, end-effector issues, and control faults) or \textit{man-made} (e.g., design flaws and user mistakes)~\cite{honig2018understanding, carlson2004follow}, and further characterized by \textit{repairability} (field vs. non-field) and \textit{impact} (terminal vs. non-terminal)~\cite{carlson2004follow}. Extensions include Guo et al.'s classification into system, operational, design, and safeguarding errors~\cite{guo2024errors}. Most existing approaches treat failure primarily as a problem of task incompletion, arising from suboptimal planning or policy decisions. For example, the RoboFail dataset~\cite{liu2023reflect} focuses on reinforcement learning policies and defines failure as the inability to successfully complete a manipulation task. While this framework is valuable for evaluating policy robustness, it largely overlooks the critical question of what happens when failures occur during execution due to hardware faults, unexpected dynamics, or other low-level disturbances that are impossible to predict or prevent in advance.

\textbf{Safety Filtering for Robotics} represents a control-theoretic approach designed to ensure system safety independently of task-driven base policies. These methods operate by monitoring policies and intervening with safe control actions when systems approach unsafe states, including control barrier functions~\cite{ames2019control, wabersich2022predictive, gupta2025t}, HJ reachability~\cite{margellos2011hamilton, mitchell2005time}, and model-predictive shielding. However, traditional safety filtering approaches primarily focus on \textit{preventing} failures through collision avoidance and constraint satisfaction. While prevention is crucial, these methods become insufficient when dealing with sudden, unpredictable hardware failures—such as joint degradation or actuator malfunctions—that cannot be easily detected or avoided in real-time. In household environments, the consequences of such failures extend far beyond direct collisions, potentially causing spilled liquids, dropped fragile objects, or cascading hazards that endanger both surroundings and humans.

\textbf{Fault Injection for Risk Analysis} has emerged as a complementary approach, focusing on understanding system behavior under various failure conditions. Yuliang et al.~\cite{ma2023case} presented a ROS-based failure injection framework using six different failure parameters and randomization capabilities. Their approach targets sensor signal faults (bias and noise) injected into position signals, demonstrating that fault phase (planning vs. execution) critically determines failure outcomes. Christensen et al.~\cite{christensen2008fault} investigated hardware fault detection using back-propagation neural networks trained on normal and faulty operational data, demonstrating robust fault identification with low false positives. While these approaches provide valuable insights into fault detection and system vulnerabilities, they primarily focus on \textit{identifying} when failures occur rather than \textit{quantifying the consequences} of failures that cannot be prevented or detected in time.

This work shifts the focus from preventing failures to \textit{assessing their impact}. When robot joints degrade suddenly or hardware fails unexpectedly—scenarios where detection or avoidance is impossible—the critical question is: what are the potential consequences at a given moment? We introduce a framework to estimate failure impacts across robot states and actions, enabling robots to make decisions that account for the \textit{severity} of failures and to choose actions that mitigate consequences when failures are unavoidable.

%% file: content/failure_mitigation.tex
\section{Problem Formulation}
\label{sec::failure_mitigation_planning}



\input{content/problem_formulation}

%% file: content/problem_formulation.tex
\begin{figure}[h]
    \centering
    \includegraphics[width=0.45\textwidth]{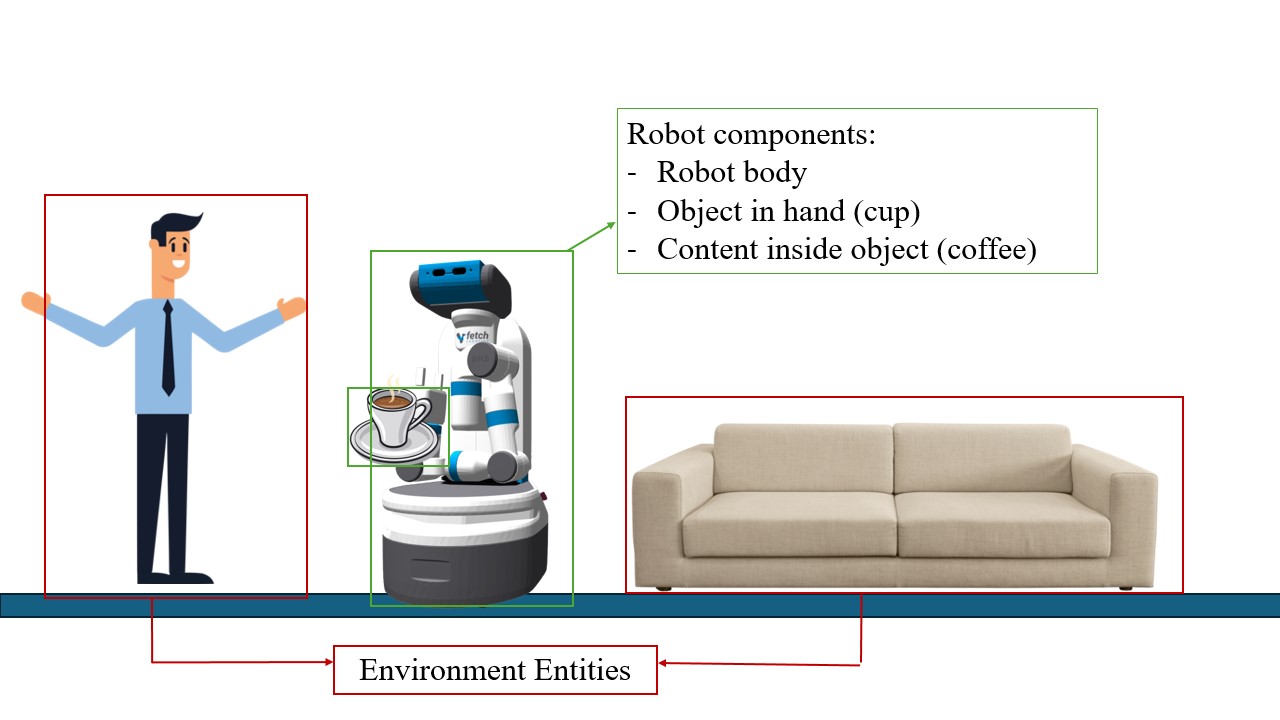} 
    \caption{Illustration of the components within the robot's workspace.}
    \label{fig:formulation_visualization}
\end{figure}

A service robot operating in human environments must minimize risks to people and objects, particularly during inevitable failures. For instance, a robot carrying a bowl of hot soup must navigate around furniture and people, as even a minor misstep could spill hot liquid and cause injury or damage. Failure-aware planning should not only enable smooth task execution but also explicitly account for the potential consequences of failures, integrating safety considerations directly into decision-making.

\subsection{Robot Components and Environment Entities}
We focus on robots operating in a household environment, representing tasks that may require both navigation and manipulation. The robot, denoted as \( R \), is divided into three primary components that can pose hazards: the robot body (\( r_1 \)), the object being carried (\( r_2 \)), and, if the object is a container, its contents (\( r_3 \)). Each component \( r_j \in R \) may present distinct risks when interacting with environmental entities. An entity \( e_i \in E \), static or dynamic, represents any object or agent in the environment that could be affected by the robot. Figure~\ref{fig:formulation_visualization} illustrates the division of the robot's workspace into components and surrounding entities.

\subsection{Interaction between Robot Components and Entities}
We define an \textbf{\textit{interaction}} as an accident following a failure that results in damage or harm to an environmental entity. At time step \( t \), the probability of an interaction is denoted as
\[
P_t(x_t, e_i, r_j \mid \text{Failure} = \text{True}),
\] 
representing the likelihood that a robot component \( r_j \) interacts with an environmental entity \( e_i \) at robot state \( x_t \) when a failure occurs.  

Since failures are assumed inevitable and undetectable, our framework does not attempt to minimize—or explicitly reason about—the probability of failure itself. Instead, it focuses on minimizing the impact of failures on the environment. Importantly, \( P_t \) captures the likelihood that a failure will negatively affect a specific entity, rather than the overall probability of failure.

\subsection{Severity of Interaction}
Not all interactions carry equal consequences. For example, spilling hot coffee is more hazardous than spilling iced water, even if the probabilities of interaction are similar. To capture this, we define a \textbf{\textit{severity}} factor \( S(e_i, r_j) \), quantifying the potential impact of an interaction between an environmental entity \( e_i \) and a robot component \( r_j \) during a failure. This factor enables the robot to adjust its behavior according to the potential severity of interactions.

\subsection{Optimization Objective}
We define the impact of a failure as the product of the probability of a harmful interaction and its severity. Using this, the motion planner minimizes the following objective:

\begin{equation}
\begin{gathered}
\underset{x_1,..., x_T}{\text{argmin}} \sum_{t=1}^T 
\Bigg[  V(x_t, x_{t-1}) \\
 + w \cdot\sum_{r_j \in R} \sum_{e_i \in E} P_{t}(x_t, e_i, r_j|F) \cdot S(e_i, r_j) 
\Bigg],
\end{gathered}
\label{eq:safe_objective}
\end{equation}
where $V(x_t, x_{t-1})$ is the cost of transitioning from state $x_{t-1}$ to $x_t$, and $w$ is a weighting parameter that balances motion efficiency against failure impact considerations.

This formulation allows the robot to evaluate the risk associated with its states and actions, balancing operational efficiency with safety. By considering both the likelihood of causing harm and the severity of potential consequences, the robot can make informed, risk-averse decisions that minimize the impact of inevitable failures.

%% file: content/failbench.tex
\section{FailBench}
\label{sec::failbench}

\label{sec::failbench}

\begin{table*}[!t]
\centering
\caption{Failure Injector for \texttt{FailBench}}
\label{tab:failure_types}
\begin{tabular}{cccc}
\toprule
& \textbf{Failure Type} & \textbf{Description} & \textbf{Onset} \\
\midrule
\multirow{7}{*}{\textbf{Actuator Failures}} & Complete Shutdown & Total loss of joint control authority & Instant \\
& Loose Joint & Reduced damping and control & Gradual \\
& Limited ROM & Restricted range of motion & Instant \\
& Partial Degradation & Reduced control authority & Gradual \\
& Gradual Degradation & Slow deterioration of performance & Gradual \\
& Stuck Position & Joint locked at current position & Instant \\
& Increased Friction & Higher resistance in joint motion & Gradual \\
\midrule
\multirow{6}{*}{\textbf{Sensor Failures}} & Noisy Sensor & Added measurement noise & Gradual \\
& Bias Drift & Accumulating systematic errors & Gradual \\
& Control Delay & Network/computational latency & Gradual \\
& Sensor Dropout & Intermittent loss of measurements & Instant \\
& Scale Factor Error & Incorrect measurement scaling & Gradual \\
& Dead Zone & No response in measurement range & Instant \\
\midrule
\multirow{3}{*}{\textbf{End-Effector Failures}} & Gripper Failure & End-effector malfunction & Instant \\
& Weak Grip & Reduced grasping force & Gradual \\
& Stuck Open/Closed & Gripper locked in position & Instant \\
\midrule
\multirow{3}{*}{\textbf{Power System Failures}} & Battery Degradation & Reduced available power & Gradual \\
& Voltage Drop & Insufficient power delivery & Gradual \\
& Power Loss & Complete system shutdown & Instant \\
\bottomrule
\end{tabular}
\vspace{0.2cm}
\end{table*}
\label{tab:failure_types}

We introduce \texttt{FailBench}, a MuJoCo~\cite{todorov2012mujoco} simulation framework for studying how robots interact with their environments under sudden, unexpected failures. \texttt{FailBench} enables systematic analysis and data generation of failure-induced behaviors, providing insights for developing proactively safe motion plans or robust learned policies. It also provides a standardized benchmark for evaluating, comparing, and developing robot planning and control strategies under various failure conditions, ultimately enhancing safety and resilience in real-world deployments.

\subsection{Simulation}
\texttt{FailBench} leverages MuJoCo as its core simulator, offering high-fidelity physics essential for accurately modeling failure behaviors. To facilitate perception-driven analyses, it provides interfaces to access ground-truth states, contact forces, and other environment metrics. Our framework incorporates scenes and object meshes curated from prior work~\cite{robocasa, robosuite}, and supports a variety of robot embodiments such as Clearpath Jackal, Fetch mobile manipulator, and Franka Panda Arm. Unlike existing platforms that primarily target at providing training environments for learning-based approaches, \texttt{FailBench} emphasizes failure-aware planning and policy evaluation.

\subsection{Motion Planning}
\subsubsection{Navigation} 
Our framework provides an interface to generate floor maps for high-level path planning, supporting both search-based algorithms such as A* and Dijkstra's, as well as sampling-based planners. For low-level control of wheel-based robots, we implement the Dynamic Window Approach~\cite{dwa} and differential-drive controllers.

\subsubsection{Manipulation}
We leverage Mink~\cite{Zakka_Mink_Python_inverse_2025}, a Python library for differential inverse kinematics built on MuJoCo. Our framework supports arm planning in both joint-space and task-space, integrating a variety of motion planners:  
\begin{itemize}  
    \item Sampling-based planners: PRM (Probabilistic Roadmaps)\cite{prm}, RRT (Rapidly-exploring Random Trees)\cite{rrt}, RRT*\cite{rrtstar}, RRTConnect\cite{rrtconnect}, and TRRT (Transition-based RRT)\cite{rrtconnect}, and
    \item Optimization-based planners: STOMP (Stochastic Trajectory Optimization for Motion Planning)\cite{stomp} and CHOMP (Covariant Hamiltonian Optimization for Motion Planning)\cite{chomp}.
\end{itemize}

This setup enables researchers to leverage, evaluate, and compare a wide range of navigation and manipulation strategies within the same simulation platform. We plan to continue expanding the library of planners in future releases.

\subsection{Failure Injector}  

A key feature of \texttt{FailBench} is its failure injector, which introduces controlled perturbations to the robot and its environment. Injecting failures enables systematic evaluation of robot safety across a variety of failure modes, including sensor noise, hardware malfunctions, and performance degradation. By simulating such scenarios, \texttt{FailBench} helps researchers assess and improve the robustness of both motion plans and learned policies. Currently, we support actuator failures such as \textit{complete shutdown} and \textit{loose joint}, sensor failures including \textit{noisy sensor} and \textit{bias drift}, and end-effector failures such as \textit{gripper failure}. More failures will be expanded in future work based on the community's feedback. 


\subsubsection{Overview}
Our failure injection framework provides a comprehensive taxonomy of three primary categories: actuator, sensor, and end-effector failures (Table~\ref{tab:failure_types}). Each failure type is characterized by its onset pattern (instant or gradual), failure degrees, and configurable parameters that control the failure's magnitude and duration. Instant failures occur immediately upon injection, while gradual failures develop or persist over time during execution.

The injector operates by modifying the underlying MuJoCo simulation parameters in real-time, enabling realistic physical simulation of robot malfunctions. Unlike approaches that only add noise to control signals, our framework directly alters actuator gains, joint constraints, sensor feedback, and mechanical properties to accurately represent the physics of actual hardware failures. This approach ensures that failure propagation and system responses mirror real-world behaviors for maximum failure fidelity.

Failures can be injected at any point during trajectory execution, with precise timing control and the ability to affect single joints, multiple joints simultaneously, or the entire system. The framework supports both deterministic failure scenarios for controlled testing and stochastic failure injection for robustness evaluation under uncertain conditions.

\subsubsection{Failure Types} \hfill 
Table~\ref{tab:failure_types} summarizes the various failure modes currently included for injection in our simulation.

\textbf{Actuator Failures} represent loss or degradation of motor control authority across multiple failure modes. \textit{Complete shutdown} removes all control from specified joints, simulating motor driver failures or emergency stops. \textit{Loose joint} failures model reduced damping and control precision from mechanical backlash or worn components. \textit{Limited ROM} restricts joint movement ranges, simulating mechanical constraints. Progressive failures include \textit{partial degradation} and \textit{gradual degradation}, which reduce control authority over time, while \textit{stuck position} locks joints at their current configuration. \textit{Increased friction} models wear-related resistance in joint mechanisms.

\textbf{Sensor Failures} affect the robot's perception of its own state and environment. \textit{Noisy sensor} failures add measurement noise to joint position feedback. \textit{Bias drift} introduces accumulating systematic errors that model sensor calibration drift over time. \textit{Control delay} simulates network latency or computational delays in the control loop. Additional failure modes include \textit{sensor dropout} for intermittent measurement loss, \textit{scale factor error} for incorrect measurement scaling due to calibration issues, and \textit{dead zone} failures where sensors become unresponsive within specific ranges.

\textbf{End-Effector Failures} specifically target manipulation capabilities. \textit{Gripper failure} encompasses complete end-effector malfunction, while \textit{weak grip} models progressive reduction in grasping force. \textit{Stuck open/closed} scenarios simulate mechanical failures that lock the gripper in fixed positions, directly impacting object manipulation success.

\textbf{Power System Failures} model energy-related degradation common in mobile robots. \textit{Battery degradation} reduces available power over time, \textit{voltage drop} causes insufficient power delivery affecting actuator performance, and \textit{power loss} simulates complete system shutdown scenarios.

Each failure type includes configurable failure degrees (minor, moderate, severe, and critical) that scale the failure's impact proportionally. Duration parameters allow for temporary failures that automatically restore after a specified time, or permanent failures that persist until explicitly restored. This flexibility enables researchers to study both transient disturbances and persistent fault conditions.

\begin{figure}[t]
    \centering
    \includegraphics[width=0.48\textwidth]{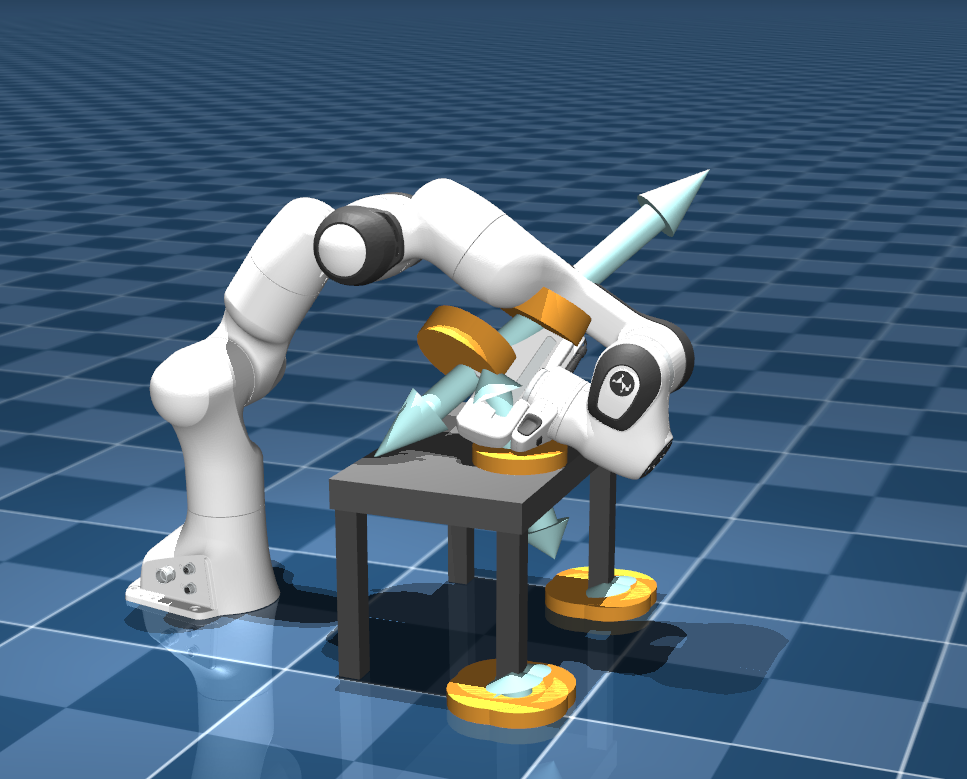}
    \caption{Visualization of contact detection after shoulder joint complete shutdown failure. The robotic arm has collided with the table, generating multiple contact points (shown in yellow) with associated force vectors, demonstrating how \texttt{FailBench} quantifies physical interaction during failure scenarios.}
    \label{fig:contact}
\end{figure}

\subsection{Data generation}

In \texttt{FailBench}, failure aftereffects are quantified through contact events, which serve as measurable proxies for the physical impact of robot failures. This approach directly aligns with the interaction framework defined in our problem formulation (Section~\ref{sec::failure_mitigation_planning}). Leveraging the MuJoCo physics engine, \texttt{FailBench} captures both contact point positions and their corresponding contact forces, providing comprehensive data for failure impact assessment. Figure~\ref{fig:contact} illustrates how contact events are detected and measured following a robot failure.

\texttt{FailBench} provides a comprehensive open-source framework with pre-configured robots, planners, environments, and object assets to streamline experimentation and accelerate research in robot safety. 

%% file: content/method.tex
\section{Failure impact analysis}
\label{sec::method}

To evaluate the failure impact formulation proposed in Section~\ref{sec::failure_mitigation_planning}, we develop computational techniques that quantify the potential impacts arising from robot–environment interactions during failure events.

\subsection{Interaction Probability Estimation}
\input{content/contact}

\subsection{Trajectory Analysis}
\input{content/traj_analysis}


%% file: content/contact.tex
Algorithm~\ref{alg:prob_interaction} outlines a general procedure for estimating interaction probabilities under joint failures in robotic manipulators. The algorithm models both individual joint failures and complete system collapse scenarios, where disabled joints cause downstream components to move freely under gravitational forces. Interaction likelihood is computed based on geometric overlap analysis in the horizontal plane between the robot's swept volume and environmental entities.

\begin{algorithm}[h]
\caption{Generalized Probability of Interaction Computation}
\label{alg:prob_interaction}
\begin{algorithmic}[1]
\Require Robot component \(r_j\), environment entity \(e_i\), robot state \(x_t\)

\State Compute \(b_{r_j} \in \mathbb{R}^{8\times3} \), the axis-aligned bounding box corners of \(r_j\) in global coordinates
\State Extract \(\min(b_{r_j}) \in \mathbb{R}^{3}, \max(b_{r_j}) \in \mathbb{R}^{3} \), the minimum and maximum coordinate values over the 8 corners and compute projected area on the \(X\text{-}Y\) plane, \(A_{r_j}\)

\State Compute \(b_{e_i} \in \mathbb{R}^{8\times3} \) 
\State Extract \(\min(b_{e_i}) \in \mathbb{R}^{3}, \max(b_{e_i}) \in \mathbb{R}^{3} \) and compute \(A_{e_i}\)

\State Compute overlap area \(A_{\text{overlap}_{i,j}}\) in \(X\text{-}Y\) plane :

\vspace{0.5ex}
\Statex \hspace{\algorithmicindent} 5a: \(
\text{min\_maxs}_{i,j} = \min( \max(b_{r_j}) , \max(b_{e_i})) \)

\vspace{0.5ex}
\Statex \hspace{\algorithmicindent} 5b: \(
\text{max\_mins}_{i,j} = \max(\min(b_{r_j}), \min(b_{e_i})) \)

\vspace{0.5ex}
\Statex \hspace{\algorithmicindent} 5c: \(\text{ax\_ov}_{i,j} = \max(\text{min\_maxs}_{i,j} - \text{max\_mins}_{i,j}, 0) \)

\vspace{0.5ex}
\Statex \hspace{\algorithmicindent} 5d:
\(A_{\text{overlap}_{i,j}} = \prod^2_{a=1}\text{ax\_ov}_{i,j}[a]\)
\vspace{0.5ex}
\State Compute probability:
\[
P_t = \max\!\left(\frac{A_{\text{overlap}_{i,j}}}{A_{r_j}}, \frac{A_{\text{overlap}_{i,j}}}{A_{e_i}}\right)
\]

\If{\(\min(b_{r_j})[3] > \max({b_{e_i})[3]}\) \textbf{or} \(\text{ax\_ov}_{i,j}[3] > 0\) }
    \State \Return \(P_t\)
\Else
    \State \Return 0
\EndIf

\end{algorithmic}
\end{algorithm}

In MuJoCo, rigid bodies are represented as collections of \textit{geoms}, so Algorithm~\ref{alg:prob_interaction} is applied at the geom level. For a robot component \(r_j\), let \(\{g^j_k\}_{k=1}^{N_j}\) denote its associated geoms, where \(N_j\) is the number of geoms. Similarly, \(\{g^i_l\}_{l=1}^{N_i}\) denotes the geoms of environment entity \(e_i\).  

Steps 1–4 of the algorithm are computed for each geom: for each \(g^j_k\) and \(g^i_l\), we calculate the 8 corners of their axis-aligned bounding boxes, \(b_{g^j_k}\) and \(b_{g^i_l}\), extract minimum and maximum coordinates, and compute the projected area on the \(X\text{-}Y\) plane, \(A_{g^j_k}\) and \(A_{g^i_l}\).  

In Step 5, the overlap area between geom pairs \((g^j_k, g^i_l)\) is computed in the \(X\text{--}Y\) plane. Specifically: (5a) finds the smaller of the two maximum coordinates, (5b) finds the larger of the two minimum coordinates, (5c) computes the difference along each axis, treating negative values as zero to get \(\text{ax\_ov}_{l,k}\), and (5d) multiplies these differences along the \(X\text{-}Y\) axis to obtain the total overlap area \(A_{\text{overlap}_{g^i_l, g^j_k}}\).  

Body-level areas are then obtained by summing over geoms:  
\[
A_{r_j} = \sum_{k=1}^{N_j} A_{g^j_k}, \quad
A_{e_i} = \sum_{l=1}^{N_i} A_{g^i_l},
\] 
\[
A_{\text{overlap}_{i,j}} = \sum_{l=1}^{N_i}\sum_{k=1}^{N_j} A_{\text{overlap}_{g^i_l, g^j_k}}.
\]

Finally, the probability of interaction between \(r_j\) and \(e_i\) is computed using Step 6. Step 7 ensures that \(e_i\) is below \(r_j\) along the \(Z\) axis by comparing minimum and maximum \(Z\) coordinates over all geoms.


While Algorithm~\ref{alg:prob_interaction} is general and capable of handling arbitrary failure types (including joint or actuator failures), in this paper we focus exclusively on \textit{object drop failures}. That is, we simulate the scenario where the robot slips the object it is carrying, without considering full manipulator collapse. This simplifies the experiment while still allowing us to evaluate the impact of failures on the environment. 

%% file: content/traj_analysis.tex
We evaluate our failure impact framework on four trajectories for a tabletop pick-and-place task in \texttt{FailBench} using the \textit{Franka Emika Panda} robot arm.

\subsubsection{Experimental Setup}

\begin{figure}[t]
    \centering
    \includegraphics[width=0.48\textwidth]{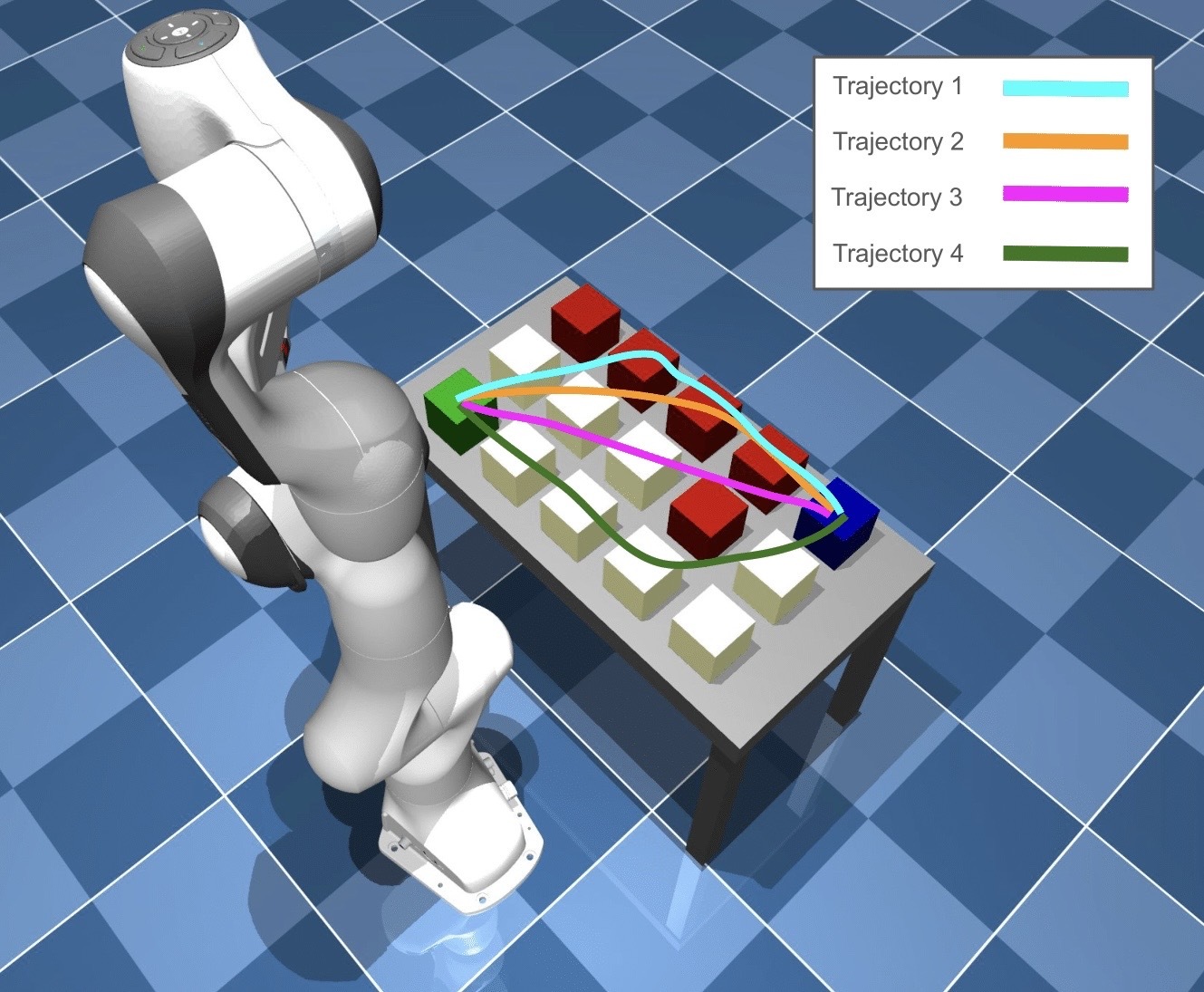}
    \caption{Trajectories carrying an object during pick-and-place execution. Trajectory 1 (light blue), 2 (orange), 3 (pink), and 4 (green) tradeoffs safety and efficiency.}
    \label{fig:4traj}
\end{figure}

The manipulation task requires the robot to pick up an object (green) from the table and place it on top of another object (blue), as shown in Figure~\ref{fig:4traj}. We sample intermediate waypoints and employ RRT-Connect to generate feasible trajectories for the Franka Emika Panda robot arm. To assess the risk of object dropping during trajectory execution, we evaluate interaction probabilities using Algorithm~\ref{alg:prob_interaction}. Object severity values are predefined in this analysis, with red objects representing high severity entities (value 10) and white objects indicating standard severity levels (value 2).

\begin{figure}[t]
    \centering
    \includegraphics[width=0.48\textwidth]{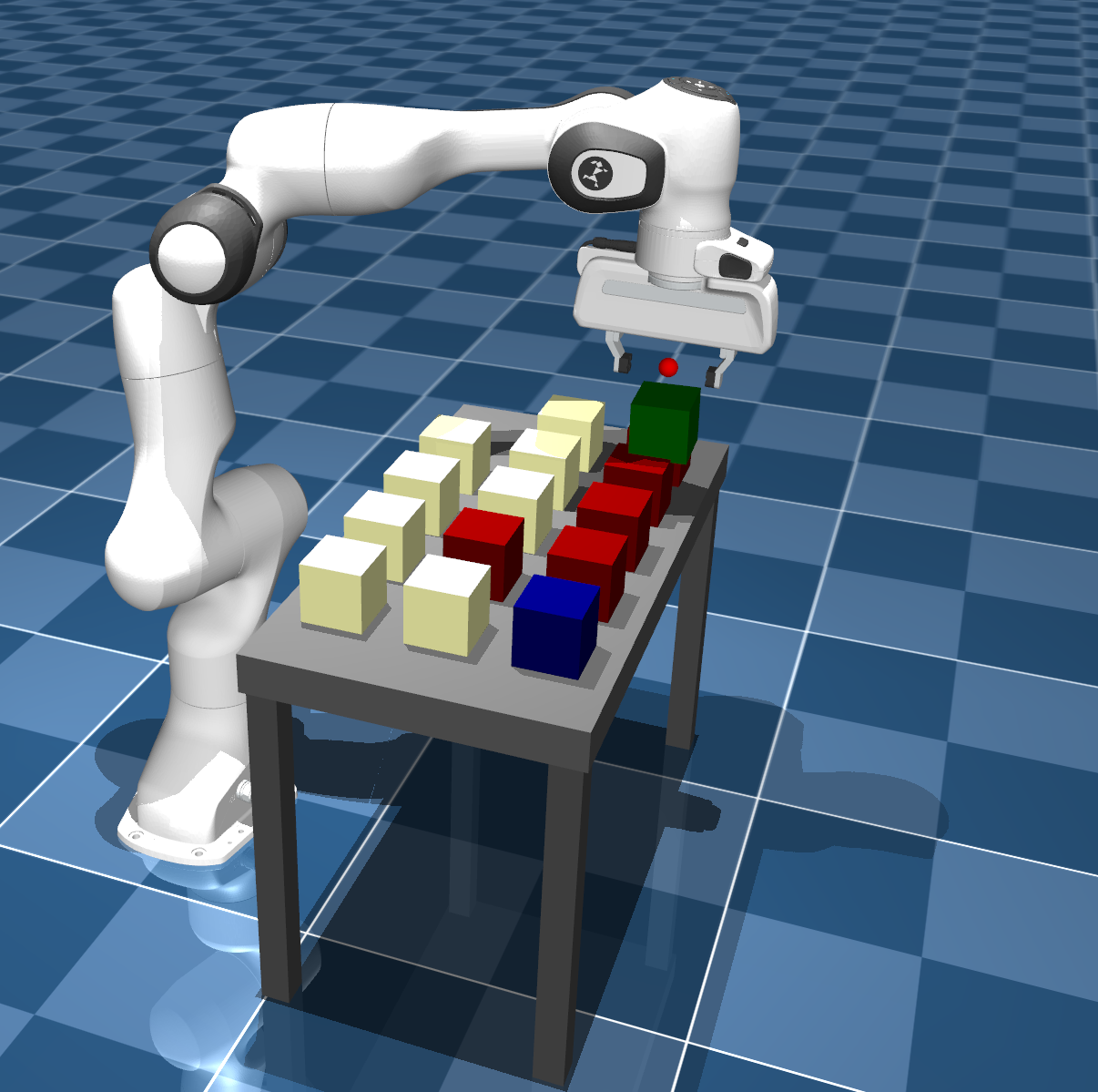}
    \caption{Visualization of object drop failure during pick-and-place execution. The robot has released the green object mid-trajectory, demonstrating the type of failure scenario in our analysis.}
    \label{fig:drop}
\end{figure}

For each trajectory, we compute the motion cost as the cumulative Euclidean distance between consecutive configurations, $\sum_{i=1}^{N-1} \|q_i - q_{i+1}\|_2$, where $\|\cdot\|_2$ denotes the Euclidean norm and $N$ is the number of waypoints. Safety cost represents the cumulative expected impact across potential drop interactions during failure scenarios, as defined in our optimization objective (Eq.~\eqref{eq:safe_objective}, with weight $w$ set to 1). To validate our framework, we simulate each trajectory 60 times with a 25\% probability of failure occurrence per trajectory execution, as shown in Figure~\ref{fig:drop}. When a failure occurs, the timestep for object dropping is randomly sampled from the trajectory duration. The observed safety (OBS) represents the average measured safety cost across all simulation runs for each trajectory.

\subsubsection{Analysis}

Table~\ref{tab:analysis_table} reveals distinct trade-offs between motion efficiency and safety across the four evaluated trajectories. Trajectory 4 (green) achieves the lowest safety cost (0.236) but incurs the highest motion cost (1.958), resulting in the highest total cost (2.194). Conversely, trajectory 2 (orange) provides optimal balance, achieving both the lowest motion cost (1.280) and total cost (1.71) despite higher safety cost than trajectories 3 (pink) and 4 (green). Trajectory 1 (light blue) demonstrates suboptimal performance with the second-highest motion cost (1.567) and total cost (1.914), while trajectory 3 (pink) serves as a reasonable middle-ground solution with moderate performance across all metrics.

The observed safety (OBS) results from simulation validation generally align with our theoretical safety cost predictions, validating the framework's predictive capability. Trajectory 4 demonstrates both the lowest theoretical safety cost (0.236) and observed safety (0.27), confirming its superior safety performance. Similarly, trajectory 3 shows low observed safety (0.4) consistent with its moderate theoretical safety cost (0.363). However, trajectory 2 exhibits higher observed safety (5.6) than expected from its theoretical cost (0.430), while trajectory 1 shows intermediate observed safety (3.33) despite having a lower theoretical safety cost (0.347) than trajectory 2. These discrepancies highlight the complexity of real failure scenarios and suggest opportunities for refining the interaction probability models.

This analysis demonstrates the practical utility of our failure impact assessment framework for evaluating trajectory safety in the presence of inevitable robot failures. While traditional planners optimize solely for geometric objectives, our framework reveals failure impact costs that could inform future failure-aware planner design. Such quantitative insights could guide the development of planners that explicitly mitigate failure consequences while balancing efficiency objectives. The framework's ability to assess different trajectory candidates provides a foundation for integrating failure impact mitigation into motion planning algorithms, enabling principled decisions that minimize harm when failures occur based on application requirements and environmental vulnerability.

\begin{table}[h]
\centering
\caption{Trajectory comparison showing theoretical safety predictions versus empirical validation through simulation.}
\label{tab:analysis_table}
\begin{tabular}{c c c c c}
\toprule
Trajectory                          & Motion Cost       & Safety Cost       & Total Cost        & OBS \\
\midrule
1 (light blue)                      & 1.567             & 0.347             & 1.914             & 3.33 \\
2 (orange)                          & \textbf{1.280}    & 0.430             & \textbf{1.71}     & 5.6 \\
3 (pink)                            & 1.359             & 0.363             & 1.722             & 0.4 \\
4 (green)                           & 1.958             & \textbf{0.236}    & 2.194             & \textbf{0.27} \\
\bottomrule
\end{tabular}
\end{table}

%% file: content/future.tex
\section{Future work}
\label{sec::future}

While we demonstrate the potential of our safety mitigation framework and \texttt{FailBench} to evaluate robot safety through our proposed failure impact assessment, integrating such evaluation capabilities into practical planning systems presents several key challenges for future work.

\textbf{Environmental Knowledge Requirements.} Our framework assumes complete environmental knowledge, including object identity, material properties, and vulnerability characteristics that determine failure impact severity. However, such comprehensive prior knowledge is rarely available in practice. Visual-Language Models (VLMs) offer a promising solution by automatically estimating interaction severity from visual input. VLMs can perceive and annotate unlabeled 3D objects with semantic descriptions including material properties, fragility, and estimated significance~\cite{kabra2024leveraging}. This visual grounding enables more reliable assessment than textual descriptions alone, as VLMs observe actual object appearance, size, and context. Future work could leverage VLMs to automatically infer vulnerability parameters, enabling practical deployment of failure impact assessment.

\textbf{Failure Interaction Modeling.} Our interaction probability model relies on geometric analysis and lacks the complexity needed for realistic failure impact prediction in dynamic environments. Addressing this limitation requires accurate modeling of robot failure behaviors, physical dynamics of manipulated objects, and interactions with dynamic agents~\cite{rudenko2020human}. Since \texttt{FailBench} enables systematic simulation of post-failure states, future work can leverage this capability to collect comprehensive contact data between robot components and environment entities, enabling learning of probability distributions using neural networks. Physical world models~\cite{zhou2024dino} present another promising direction for predicting failure outcomes and component interactions. Recent advances in latent-space Hamilton-Jacobi reachability~\cite{nakamura2025generalizing} offer valuable insights for learning failure interaction representations directly from raw observations.

\textbf{Integration with Motion Planning.} A critical challenge lies in effectively incorporating failure impact assessment into motion planning algorithms. Traditional sampling-based planners excel at finding feasible paths but struggle with complex cost landscapes inherent in failure-aware planning. Optimization-based approaches face convergence challenges due to irregular cost surfaces created by discontinuous failure events. Future work should explore planning in latent spaces to achieve smoother cost representations and improved computational efficiency, enabling practical deployment of failure-aware trajectory generation.

\textbf{Task-Level Failure Awareness.} Ultimately, intelligent systems should consider failure impact not only during trajectory generation but throughout high-level task planning. Systems should prioritize actions that mitigate failure impact across entire task sequences~\cite{anticipatory_tamp}. For example, when transporting fragile objects, optimal strategies might first acquire protective equipment to minimize damage risk. Extending our evaluation framework to Task and Motion Planning (TAMP)~\cite{itamp} represents a natural progression toward comprehensive failure-aware robotics, where our benchmarking capabilities can guide the development of safer autonomous systems.

%% file: content/conclusions.tex
\section{Conclusions}
\label{sec::conclusions}

This paper presents a comprehensive framework for evaluating robot safety by explicitly accounting for failure consequences in robotic systems. We introduce a novel formulation that quantifies both the probability and severity of robot-environment interactions during failures, providing a systematic approach to assess how robots handle inevitable malfunctions while protecting their surroundings. Our method evaluates the impact of failures on different entities in the environment, establishing rigorous metrics for safety assessment in service robotics.

To enable systematic evaluation of failure scenarios, we also develop \texttt{FailBench}, a MuJoCo-based simulation framework that facilitates controlled study of robot-environment interactions under diverse failure modes, including sensing issues, hardware malfunctions, and actuator degradation. \texttt{FailBench} serves as a comprehensive benchmarking platform for robot safety research, featuring curated household environments, standardized motion planning implementations, diverse failure injection modes, and quantitative metrics for failure impact assessment. This framework enables researchers to systematically assess robot failure data and compare different approaches to failure handling and safety-aware robotics.


Our evaluation demonstrates the framework's capability to assess robot safety across various scenarios and planning algorithms. To integrate our framework into motion planning, we identify several challenges and pathways for advancing failure-aware robotics. Integration with Visual-Language Models for automatic scene understanding, learning-based probability models trained on \texttt{FailBench} data, and extension to Task and Motion Planning represent promising directions enabled by our benchmarking foundation.

These contributions establish a critical foundation for safer, more reliable service robots operating in dynamic household environments. By shifting focus from pure failure prevention to systematic consequence evaluation, our work provides the tools and metrics necessary for developing robots that can be rigorously assessed for their ability to gracefully handle inevitable failures while maintaining safety and user trust.